\documentclass{interact}
\usepackage{graphicx} % Required for inserting images
\usepackage{xcolor}
\usepackage[utf8]{inputenc} % For UTF-8 encoding
\usepackage[numbers]{natbib}         % For citation management
\usepackage{hyperref}       % For hyperlinks, including DOIs
\usepackage{bbm}
\usepackage{epstopdf}
\usepackage{amssymb}

\usepackage{tikz}

\title{\underline{O}pen-Set \underline{3D} \underline{S}emantic \underline{I}nstance \underline{M}aps for Vision Language Navigation - O3D-SIM}

\author{
\name{Laksh Nanwani\textsuperscript{a}\thanks{Contact - Laksh Nanwani. Email: lakshanshul@gmail.com}, Kumaraditya Gupta\textsuperscript{a*}, Aditya Mathur\textsuperscript{a*}, Swayam Agrawal\textsuperscript{a}, \\A.H. Abdul Hafez\textsuperscript{b}, K. Madhava Krishna\textsuperscript{a}}
\affil{\textsuperscript{a}Robotics Research Center, International Institute of Information Technology, Hyderabad, India
\\
\textsuperscript{b}Faculty of Engineering, Hasan Kalyoncu University, Turkey}
}
\begin{document}

\maketitle

% --- This block draws the notice ---
\begin{tikzpicture}[remember picture, overlay]
  \node at (current page.north) [
    anchor=north, 
    yshift=-1.0in, % Adjust this value to move it up or down
    text width=\textwidth, 
    align=center
  ]
  {
    \textit{This article is published in Taylor \& Francis Advanced Robotics. \\
    DOI - \href{https://doi.org/10.1080/01691864.2024.2395926}{10.1080/01691864.2024.2395926}}
  };
\end{tikzpicture}\
% ---------------------------------

\def\thefootnote{*}\footnotetext{These authors contributed equally to this work}

\begin{abstract}
Humans excel at forming mental maps of their surroundings, equipping them to understand object relationships and navigate based on language queries. Our previous work SI Maps~\cite{Nanwani_2023} showed that having instance-level information and the semantic understanding of an environment helps significantly improve performance for language-guided tasks. We extend this instance-level approach to 3D while increasing the pipeline's robustness and improving quantitative and qualitative results. Our method leverages foundational models for object recognition, image segmentation, and feature extraction. We propose a representation that results in a 3D point cloud map with instance-level embeddings, which bring in the semantic understanding that natural language commands can query. Quantitatively, the work improves upon the success rate of language-guided tasks. At the same time, we qualitatively observe the ability to identify instances more clearly and leverage the foundational models and language and image-aligned embeddings to identify objects that, otherwise, a closed-set approach wouldn't be able to identify.
\\
Project Page - \url{https://smart-wheelchair-rrc.github.io/o3d-sim-webpage}

\end{abstract}


%%%%%%%%%%%%%%%%%%%\section{Introduction}
%%%%%%%%%%%%%%%%%%%%%%%%%%%%%%%%%%%%%%%%%%%%%%%%%%%%%%%%%%%%%%%%%%%%%%%%%%%%%%%%
\section{Introduction} 
Vision-Language Navigation(VLN) research has recently seen significant progress, enabling robots to navigate environments using natural language instructions~\cite{huang23vlmaps,park2023visual}. A major challenge in this field is grounding language descriptions onto real-world visual observations, especially for tasks requiring identifying specific object instances and spatial reasoning~\cite{conceptgraphs}. There is an increasing interest in building semantic spatial maps representing the environment and its objects~\cite{placed2023survey}. Works like VLMaps~\cite{huang23vlmaps} and NLMap~\cite{chen2022nlmapsaycan} leverage pre-trained vision-language models to construct semantic spatial maps (SSMs) without manual labelling. However, these methods cannot differentiate between multiple instances of the same object, limiting their utility for instance-specific queries.

Our previous work~\cite{Nanwani_2023} proposes a memory-efficient mechanism for creating a 2D semantic spatial representation of the environment with instance-level semantics directly applicable to robots navigating in real-world scenes.  It was shown that \textbf{Semantic Instance Maps} (\textit{SI Maps}) are computationally efficient to construct and allow for a broad range of complex and realistic commands that elude prior works. However, the map built in this previous work is a top-down view 2D map based on a closed set and assumes that the Large Language Model (LLM) knows all object categories in advance to map language queries to a specific object class. The 2D maps, though sufficient for a good range of tasks, limit performance when larger ones can obscure smaller objects.

Building upon~\cite{Nanwani_2023}, we introduce a novel approach for Semantic Instance Maps. Our work introduces \textbf{Open-set 3D Semantic Instance Maps} (\textit{O3D-SIM}), addressing the limitation of traditional closed-set methods~\cite{cheng2021mask2former, He_2017_ICCV}, which assume only predefined set objects will be encountered during operation. In contrast, O3D-SIM leverages methods to identify and potentially categorize unseen objects not explicitly included in the training data. This is crucial for scenarios with diverse and unseen objects. This enables better performance and complex query handling for unseen object categories.  
We address the previously mentioned limitations by enabling instance-level object identification within the spatial representation and operating in an open-set manner, excelling in real-world scenarios.
We have achieved major improvements over our previous work, which operated in a closed-set manner. The current pipeline leverages state-of-the-art foundational models like CLIP~\cite{radford2021learning_clip} and DINO~\cite{oquab2024dinov2}. These models extract semantic features from images, allowing them to recognize objects and understand the finer details and relationships between them. For example, the DINOv2~\cite{oquab2024dinov2} model, trained on various chair images, can identify a chair in a new image and distinguish between a dining chair and an office chair.

This paper details the proposed pipeline for creating the 3D map and evaluates its effectiveness in VLN tasks for both simulation and real-world data. Experimental evaluations demonstrate the improvements achieved through our open-set approach, with an increase in correctly identified object instances, bringing our results closer to ground truth values even in challenging real-world data. Additionally, our pipeline achieves a higher success rate for complex language navigation queries targeting specific object instances, including those unseen during training, such as mannequins or bottles.

The contributions of our work can be summarized as follows:
\begin{enumerate}
    \item Extending the closed-set 2D instance-level approach from our previous work~\cite{Nanwani_2023} to an open-set 3D Semantic Instance Map using state-of-the-art image segmentation, open-set image-language-aligned embeddings, and hierarchical clustering in 3D. The resulting map consists of a 3D point cloud with instance-level embeddings, enhancing the semantic understanding.
    \item Validating the effectiveness of the 3D map approach to VLN tasks through both qualitative and quantitative experiments.
\end{enumerate}

\begin{figure}
    \centering
    \includegraphics[width=1\linewidth]{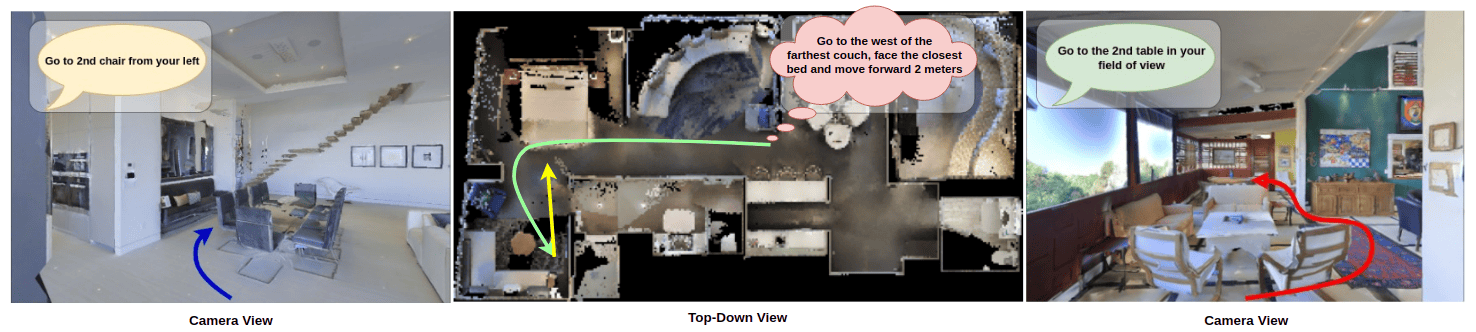}
   \caption{We carry out complex instance-specific goal navigation in object-rich environments. These language queries refer to individual instances based on spatial and viewpoint configuration concerning other objects of the same type while preserving the navigation performance on standard language queries.}
  \label{fig:figure1}
  \vspace{-1em}
\end{figure}

The remainder of this paper is organized as follows. Section~\ref{sec:related_works} reviews and analyzes recent literature, providing the necessary background on semantic scene understanding, 3D scene reconstruction, and VLN to the reader. Section~\ref{sec:method} outlines the methodology behind our proposed 3D map. Section~\ref{sec:experiments} discusses the experimental evaluation of our proposed 3D map's effectiveness in achieving complex language queries as shown in Figure \ref{fig:figure1}. Finally, concluding remarks and directions for future work are presented in Section~\ref{sec:futureworkconclusion}.

\section{Related Works}\label{sec:related_works}
\subsection{Vision-and-Language Navigation:} 
Vision-and-Language Navigation (VLN) has recently gained much traction because of its potential to improve autonomous navigation by combining human natural language understanding and visual perception~\cite{huang23vlmaps, chen2022nlmapsaycan, chang2023goat, Anderson_2018_CVPR}. The development of foundation models like CLIP~\cite{radford2021learning_clip}, which combines image and text data to learn the rich representations of the environment, has spurred considerable progress in vision and language understanding. The multi-modal nature of these foundation models allows them to comprehend concepts in both text and image and even connect concepts between the two modalities. There have been increasing advancements to address the gaps faced by previous methods for VLN, like navigation efficiency to spatial goals specified by language commands and zero-shot spatial goal navigation given unseen language instructions~\cite{huang23vlmaps}.

A major challenge in VLN is interpreting language instructions in unfamiliar environments. A significant limitation of previous studies in this domain is their handling of action errors. Suppose a robot agent makes an incorrect action. In that case, it risks failing to reach its destination or exploring unnecessary areas, leading to increased computational demands and possibly entering a state from which recovery is unfeasible. State-of-the-art VLN methodologies employ diverse strategies to excel in such scenarios. Some methods adopt a specialized pre-training and fine-tuning approach designed explicitly for VLN tasks, utilizing transformer-based architectures~\cite{10.5555/3295222.3295349}. These strategies often involve using image-text-action triplets in a self-supervised learning context~\cite{9156554}. Other approaches refine the pre-training process to enhance VLN task performance, for instance, by emphasizing the learning of spatiotemporal visual-textual relationships to better utilize past observations for future action prediction~\cite{9880046, 2023metaexplore, 10160259}.

In this work, we also address some of the limitations observed in existing end-to-end VLN methods, such as those detailed in Qi et al. \cite{Qi2021TheRT}, Huo et al. \cite{Huo2023GeoVLNLG}, and Hong et al. \cite{Hong2020VLNBERTAR}. These approaches rely on direct action estimation from multimodal inputs without constructing an environmental map, necessitating highly specific navigational instructions, which can feel unnatural and rigid in real-world settings. Furthermore, these methods modify BERT~\cite{devlin2018bert} architectures to parse language commands, thus missing out on the broader contextual understanding offered by newer larger language models (LLMs) like GPT 3.5. These models also tend to process multiple RGB images per timestep to determine actions, leading to higher computational demands during inference. In contrast, our method integrates environmental mapping, which enhances the system's spatial understanding and allows for more context-aware navigation.

Furthermore, contemporary VLN systems predominantly rely on simulations due to their dependency on panoramic views and extracting region features, which can be computationally prohibitive. In contrast, our work demonstrates our pipeline's efficiency and computational viability with real-world data, underscoring its practical applicability.

Given the recent advances in the semantic understanding of images, there has been an increasing interest in using semantic reasoning to improve exploration efficiency in novel environments and handling semantic goals specified via categories, images, and language instructions. Most of these methods are specialized to a single task, i.e. they are uni-modal.

Recent works have also been on executing tasks with lifelong learning, which means taking advantage of experience in the same environment for multi-modal navigation~\cite{chang2023goat}. One such task is that a robot must be able to reach any object specified in any way and remember object locations to return to them. The work done on these tasks also utilizes CLIP to align both image and language embeddings, where they match language goal descriptions with all instances in the environment using the cosine similarity score between their CLIP features.

Other approaches to VLN, such as~\cite{Nanwani_2023, huang23vlmaps, conceptgraphs, conceptfusion}, involve building a map of the environment along with the semantic understanding and utilising this global-level representation to achieve multimodal tasks.  These works have shown that understanding the environment on a global scale helps achieve VLN tasks easily and saves the agent from unnecessary explorations that may occur due to wrong future-action predictions. Our approach falls under the same category.

\subsection{Semantic Scene Understanding and Instance Segmentation}

Our work builds upon recent advancements in semantic and instance segmentation of 3D scenes. This domain has been thoroughly explored using closed-set vocabulary methods, including our prior work \cite{Nanwani_2023}, which utilizes Mask2Former \cite{cheng2021mask2former} for image segmentation. Various studies \cite{8741085, mascaro2022volumetric, miao2023volumetric} have adopted a similar approach to achieve object segmentation, resulting in a closed-set framework. While these methods are effective, they are constrained by the limitation of predefined object categories. Our approach employs SAM \cite{kirillov2023segment} to acquire segmentation masks for open-set detection. Moreover, our methodology, distinct from many existing techniques that depend heavily on extensive pre-training or fine-tuning, integrates these models to forge a more comprehensive and adaptable 3D scene representation. This emphasizes enhanced semantic understanding and spatial awareness.

To improve the semantic understanding of the objects detected within our images, we harness detailed feature representations using two foundational models: CLIP \cite{radford2021learning_clip} and DINOv2 \cite{oquab2024dinov2}. DINOv2, a Vision Transformer trained through self-supervision, recognises pixel-level correspondences between images and captures spatial hierarchies. Compared to CLIP, DINOv2 more effectively distinguishes between two distinct instances of the same object type, which poses challenges for CLIP.

It's crucial to differentiate individual instances following the semantic identification of objects. Early methods employed a Region Proposal Network (RPN) to predict bounding boxes for these instances \cite{8237584}. Alternatively, some strategies suggest a generalized architecture for managing panoptic segmentation \cite{kirillov2019panoptic}. In our preceding approach, we utilized the segmentation model Mask2Former \cite{cheng2021mask2former}, which employs an attention mechanism to isolate object-centric features. Recent research also tackles semantic scene understanding using open vocabularies \cite{takmaz2023openmask3d}, utilizing multi-view fusion and 3D convolutions to derive dense features from an open-vocabulary embedding space for precise semantic segmentation. Our current pipeline leverages Grounding DINO \cite{liu2023grounding} to generate bounding boxes, which are then input into the Segment Anything Model (SAM) \cite{kirillov2023segment} to produce individual object masks, thus enabling instance segmentation within the scene.

\subsection{3D Scene Reconstruction}
In recent times, 3D scene reconstruction has seen significant advancements. Some recent works in this field include using a self-supervised approach for Semantic Geometry completion and appearance reconstruction from RGB-D scans such as \cite{huang2023ssr2d}, which uses 3D encoder-decoder architecture for geometry and colour. For these approaches, the focus is on generating semantic reconstruction without ground truth. Another approach is to integrate real-time 3D reconstruction with SLAM. This is done through keyframe-based techniques and has been used in recent autonomous navigation and AR use cases\cite{picard2023survey}. 
Another recent method has seen work on Neural Radiance Fields\cite{mildenhall2020nerf} for indoor spaces when utilizing structure-from-motion to understand camera-captured scenes. These NeRF models are trained for each location and are particularly good for spatial understanding. Another method is to build 3D scene graphs using open vocabulary and foundational models like CLIP to capture semantic relationships between objects and their visual representations\cite{conceptgraphs}. During reconstruction, they use the features extracted from the 3D point clouds and project them onto the embedding space learned by CLIP.

This work uses an open-set 2D instance segmentation method, as explained in the previous sections. Given an RGB-D image, we get these individual object masks from the RGB image and back-project them to 3D using the Depth image. Here, we have an instance-based approach instead of having a point-by-point computation to reconstruct, which was previously done by Concept-Fusion~\cite{conceptfusion}. This per-object feature mask extraction also helps us compute embeddings, which preserve the open-set nature of this pipeline.

%%%%%%%%%%%%%%%%%%%%%%%%%%%%%%
\section{Methodology}\label{sec:method}
  
In this section, we discuss the pipeline of our Vision-Language Navigation (VLN) method, which employs O3D-SIM. 
We begin with an overview of our proposed pipeline and then present an in-depth analysis of its constituent steps. The initial phase of our methodology involves data collection, consisting of a set of RGB-D images and extrinsic and intrinsic camera parameters,  which are outlined first. Subsequently, we move to creating the Open-set 3D Semantic Instance Map. This process is divided into two main stages: initially, we extract open-set semantic instance information from the images; following this, we utilize the gathered open-set information to organize the 3D point cloud into an open-set 3D semantic instance map. The final part of our discussion focuses on the VLN module, where we talk about its implementation and functionality.

The pipeline of the O3D-SIM creation is depicted in Figure~\ref{fig:pipeline}. The first step of the creation of the O3D-SIM, presented in Section~\ref{sec:OSfeatures}, is the extraction of the open-set semantic instance information from the RGB sequence of input images. %, see Fig.~\ref{fig:1stStage3DMap}. 
This information includes, for each object instance, the mask information and the semantic features represented by the CLIP~\cite{radford2021learning_clip} and DINO~\cite{oquab2024dinov2} embedding features. The second step, presented in Section~\ref{sec:clustering}, uses this open-set semantic instance information to cluster the input 3D point cloud into an open-set semantic 3D objects map, see Figures~\ref{fig:pipeline} and \ref{fig:clustering}. The operation is improved incrementally by applying the sequence of RGB-D images over time. 

\subsection{Data Collection}
Creating the O3D-SIM begins by capturing a sequence of RGB-D images using a posed camera, with an estimate of the extrinsic and intrinsic parameters of the environment to be mapped. The pose information associated with each image is used to transform the point clouds to a world coordinate frame. For simulations, we use the ground-truth pose associated with each image, whereas we leverage RTAB-Map\cite{labbe2019rtab} with G2O optimization~\cite{5979949} in the real world to generate these poses.

\subsection{Open-set Semantic Information from Images }\label{sec:OSfeatures}

\begin{figure} 
    \centering 
    \includegraphics[width=\linewidth]{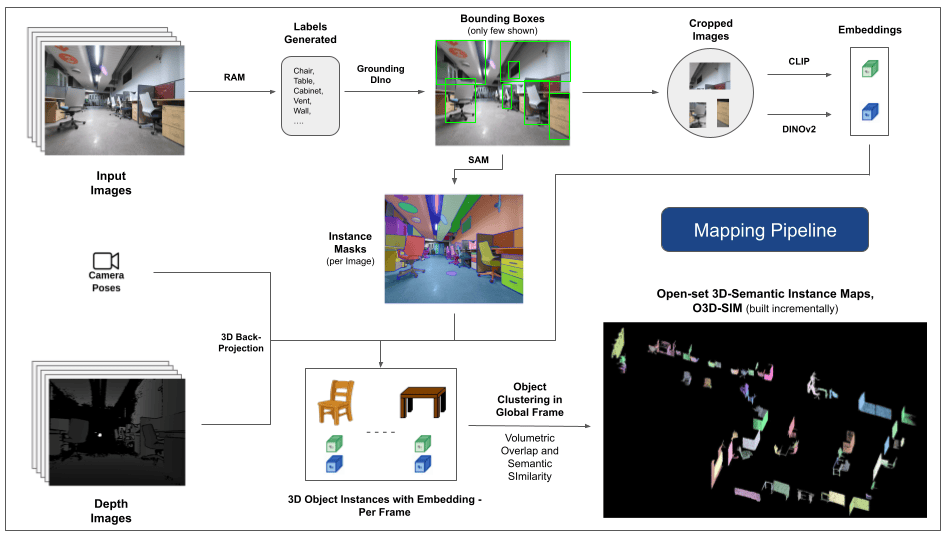}
    \caption{An overview of the proposed 3D mapping pipeline. Labels generated by the RAM model are input into Grounding DINO to generate bounding boxes for the detected labels. Subsequently, instance masks are created using the SAM model, while CLIP and DINOv2 embeddings are extracted in parallel. These masks, along with the semantic embeddings, are back-projected into 3D space to identify 3D instances. These instances are then refined using a density-based clustering algorithm to produce the O3D-SIM. Figure\ref{fig:clustering} shows how the object instance clustering works as part of O3D-SIM.}
     \label{fig:pipeline}
\end{figure}

The sequence of RGB-D images collected above is the input at this stage. Using the $n$ RGB-D images $\mathbf{I} = \{ I_1, ...,  I_n\}$, a set of object instances are extracted after processing every image in $I$. Each image $I_i$ provides us with a set $\mathbf{O_i}=\{O_{i1}, ..., O_{ij}, ... , O_{iJ_i} \}$ of object instances, where $J_i$ is the maximum of instances from the $i^{th}$ image. An object instance is represented using a mask and semantic features, i.e. CLIP~\cite{radford2021learning_clip} and DINO~\cite{oquab2024dinov2} embeddings. The flow diagram of the process of extracting the open-set semantic representation and the structure used to store the information are depicted in Figure~\ref{fig:pipeline}.

\subsubsection{Open-set Semantic and Instance Masks Detection }

The recently released Segment Anything model (SAM)~\cite{kirillov2023segment} has gained significant popularity among researchers and industrial practitioners due to its cutting-edge segmentation capabilities. However, SAM tends to produce an excessive number of segmentation masks for the same object. We adopt the Grounded-SAM~\cite{ren2024grounded} model for our methodology to address this. This process involves generating a set of masks in three stages, as depicted in Figure~\ref{fig:pipeline}. Initially, a set of text labels is created using the Recognizing Anything model (RAM)~\cite{zhang2023recognize}. Subsequently, bounding boxes corresponding to these labels are created using the Grounding DINO model~\cite{liu2023grounding}. The image and the bounding boxes are then input into SAM to generate class-agnostic segmentation masks for the objects seen in the image. We provide a detailed explanation of this approach below, which effectively mitigates the problem of over-segmentation by incorporating semantic insights from RAM and Grounding-DINO.

The RAM model)~\cite{zhang2023recognize} processes the input RGB image to produce semantic labelling of the object detected in the image. It is a robust foundational model for image tagging, showcasing remarkable zero-shot capability in accurately identifying various common categories. The output of this model associates every input image with a set of labels that describe the object categories in the image. 
The process begins with accessing the input image and converting it to the RGB colour space, then resized to fit the model's input requirements, and finally transforming it into a tensor, making it compatible with the analysis by the model. Following this, the RAM model generates labels, or tags, that describe the various objects or features present within the image. A filtration process is employed to refine the generated labels, which involves the removal of unwanted classes from these labels. Specifically, irrelevant tags such as "wall", "floor", "ceiling", and "office" are discarded, along with other predefined classes deemed unnecessary for the context of the study. Additionally, this stage allows for the augmentation of the label set with any required classes not initially detected by the RAM model. Finally, all pertinent information is aggregated into a structured format. Specifically, each image is catalogued within the img$\_$dict dictionary, which records the image's path alongside the set of generated labels, thus ensuring an accessible repository of data for subsequent analysis.

Following the tagging of the input image with generated labels, the workflow progresses by invoking the Grounding DINO model~\cite{liu2023grounding}. This model specializes in grounding textual phrases to specific regions within an image, effectively delineating target objects with bounding boxes. This process identifies and spatially localizes objects within the image, laying the groundwork for more granular analyses.
After identifying and localising objects via bounding boxes, the Segment Anything Model (SAM)~\cite{kirillov2023segment} is employed. The SAM model's primary function is to generate segmentation masks for the objects within these bounding boxes. By doing so, SAM isolates individual objects, enabling a more detailed and object-specific analysis by effectively separating the objects from their background and each other within the image.  

At this point, instances of objects have been identified, localized, and isolated. Each object is identified with various details, including the bounding box coordinates, a descriptive term for the object, the likelihood or confidence score of the object's existence expressed in logits, and the segmentation mask. Furthermore, every object is associated with CLIP and DINOv2 embedding features, details of which are elaborated in the following subsection.
 
%%%%%%%%%%%%%==========================
\subsubsection{The Semantic Embedding Extraction}
To improve our comprehension of the semantic aspects of object instances that have been segmented and masked within our images, we employ two models, CLIP~\cite{radford2021learning_clip} and DINOv2~\cite{oquab2024dinov2}, to derive the feature representations from the cropped images of each object. A model trained exclusively with CLIP achieves a robust semantic understanding of images but cannot discern depth and intricate details within those images. On the other hand, DINOv2 demonstrates superior performance in depth perception and excels at identifying nuanced pixel-level relationships across images. As a self-supervised Vision Transformer, DINOv2 can extract nuanced feature details without relying on annotated data, making it particularly effective at identifying spatial relationships and hierarchies within images. For instance, while the CLIP model might struggle to differentiate between two chairs of different colours, such as red and green, DINOv2's capabilities allow such distinctions to be made clearly. To conclude, these models capture both the semantic and visual features of the objects, which are later used for similarity comparisons in the 3D space.

A set of pre-processing steps is implemented for processing images with the DINOv2 model. These include resizing, centre cropping, converting the image to a tensor, and normalizing the cropped images delineated by the bounding boxes. The processed image is then fed into the DINOv2 model alongside labels identified by the RAM model to generate the DINOv2 embedding features. 
On the other hand, when dealing with the CLIP model, the pre-processing step involves transforming the cropped image into a tensor format compatible with CLIP, followed by the computation of embedding features. These embeddings are critical as they encapsulate the objects' visual and semantic attributes, which are crucial for a comprehensive understanding of the objects in the scene. These embeddings undergo normalization based on their L2 norm, which adjusts the feature vector to a standardized unit length. This normalization step enables consistent and fair comparisons across different images.

In the implementation phase of this stage, we iterate over each image within our data and execute the subsequent procedures:

\begin{enumerate}
    \item The image is cropped to the region of interest using the bounding box coordinates provided by the Grounding DINO model, isolating the object for detailed analysis.
    \item Generate DINOv2 and CLIP embeddings for the cropped image.
    \item Finally, the embeddings are stored back along with the masks from the previous section. 
\end{enumerate}

With these steps completed, we now possess detailed feature representations for each object, enriching our dataset for further analysis and application.

\subsection{Creating the Open-set 3D Representation}\label{sec:clustering}
\begin{figure}
    \centering
    \includegraphics[width=\linewidth]{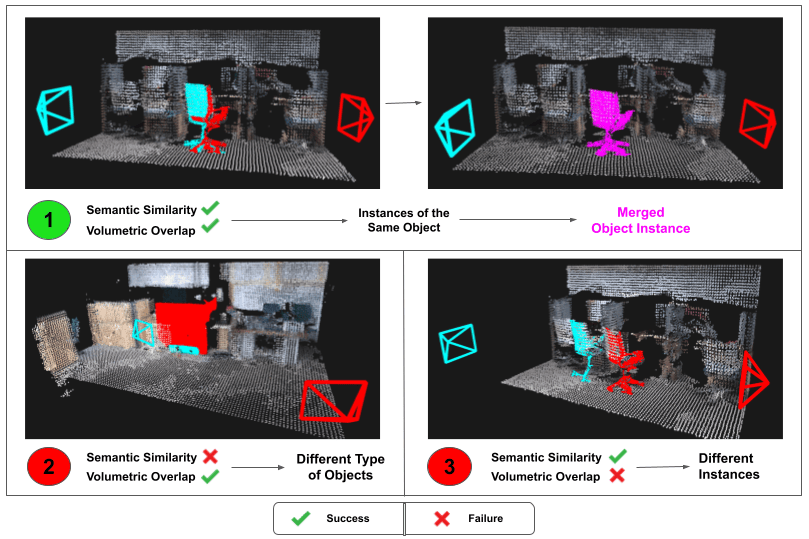}
    \caption{Clustering in 3D of object instances using semantic embeddings and volumetric overlap. Semantic similarity is verified using CLIP and DINOv2 embeddings. Volumetric overlap is calculated using 3D bounding boxes and overlap matrices. The above figure shows the clustering process between two camera poses, represented by the coloured frustum, and the corresponding object point clouds are the same colour. For the O3D-SIM pipeline, this process is repeated for each camera pose, and the objects from each frame are either merged into an existing instance or added as a new instance into the incrementally built representation. The first example shows a positive case of merging, where a comparison is made for the same instance of a chair from different poses. The chairs are merged into a single instance due to success in both semantic similarity and volumetric overlap. The second example shows a case where the table (red) and the skateboard (blue), being very close to each other, have a volumetric overlap but are not merged due to a failure in semantic similarity, creating 2 different object instances. Example 3 shows an example of two separate chairs near each other. Since they are 2 different chairs, they fail to have a volumetric overlap, creating 2 separate instances.}
    \label{fig:clustering}
\end{figure}

To complete building the O3D-SIM, we now build upon the feature embeddings extracted for each object by projecting object information to 3D space, clustering, and associating objects across multiple images to create a comprehensive 3D scene representation. The process of projecting the semantic information into the 3D space and refining the map is depicted in Figure \ref{fig:clustering}. 
%
%%%%%%%%%%%%%%%==========================
\subsubsection{The O3D-SIM Intialization}
%Initializing the scene representation 
The 3D map is initially created using a selected image, which acts as the reference frame for initialising our scene representation. This step establishes the foundational structure of our 3D scene, which is then progressively augmented with data from subsequent images to enrich the scene's complexity and detail.

The data for objects within a 3D scene are organized as nodes within a dictionary, which initially starts as empty. Objects are then identified from the initial image along with the related data that encompasses embedding features and information on their masks. For each object discerned in the image, a 3D point cloud is created using the available depth information and the object's mask. This point cloud formation involves mapping the 2D pixels into 3D space, facilitated by the camera's intrinsic parameters and depth values. Subsequently, the camera pose is utilized to align the point cloud accurately within the global coordinate system. To refine our scene representation, background filtering removes elements identified as background, such as walls or floors. These elements are excluded from further processing, particularly in the clustering stage, as they do not constitute the main focus of our scene representation.

The set of object's point clouds is processed further using DBSCAN\cite{ester1996density} clustering for representation refinement. The point cloud is downsampled via voxel grid filtering to reduce the number of points and the computational complexity while preserving the data spatial structure manageable. 

DBSCAN groups the points that are closely packed together while labelling points that lie alone in low-density regions as noise. %Fast\_DBSCAN leverages GPU via CuPy for enhanced performance, enabling efficient clustering of large point clouds crucial for 3D scene reconstruction. 
In a post-clustering step, the largest cluster typically corresponds to the main object of interest within the point cloud is identified.  This helps filter out the noise and irrelevant points, producing a cleaner representation of the object of interest.

The pose of an object in 3D space is determined by calculating the orientation of a bounding box, which offers a concise spatial representation of the object's location and size in 3D space. Subsequently, the 3D map output is initialized with an initial set of nodes, encapsulating feature embeddings, point cloud data, bounding boxes, and the count of points in the point cloud associated with each node. Each node also includes source information to facilitate tracing data origins and the linkage between nodes and their 2D image counterparts.

%%%%%%%%%%%%%%%==============================
\subsubsection{Incremental Update of the  O3D-SIM}
After initializing the scene, we update the representation with data from new images. This process ensures our 3D scene stays current and precise as additional information becomes available. It iterates over each image in the image sequence; for each new image, multi-object data is extracted, and the scene is updated.

Objects are detected for each new image, and new nodes are created like the initial image. These temporary nodes contain the 3D data for newly detected objects that must either be merged into the existing scene or added as new nodes. The similarity between newly detected and existing scene nodes is determined by combining visual similarity, derived from feature embeddings, and spatial similarity, obtained from point cloud overlap, to formulate an aggregate similarity measure. If this measure surpasses a predetermined threshold, the new detection is deemed to correspond to an existing object in the scene. Indeed, the newly detected node is either merged with an existing scene node or added as a new node.

% Volumetric Overlap of Two Pointclouds
The volumetric overlap of two point clouds is defined as the proportion of points in point cloud $pt_{i}$ that have nearest neighbours in point cloud $pt_{j}$, within a distance threshold $\delta_{nn}$. The metric can be mathematically expressed as:
\begin{equation}
\text{Volumetric overlap} = \text{nnratio}(pt_{i}, pt_{j}) = \frac{1}{|pt_{i}|} \sum_{p \in pt_{i}} \mathbbm{1}(\min_{q \in pt_{j}} \| p - q \| \leq \delta_{nn})
\end{equation}
where $\mathbbm{1}$ is the indicator function that is 1 if the condition is true and 0 otherwise, $p$ and $q$ are points in point clouds $pt_{i}$ and $pt_{j}$, respectively, and $\| p - q \|$ is the Euclidean distance between points $p$ and $q$.

% Semantic Similarity
The semantic similarity between two point clouds is calculated using the cosine similarity between the DINO features of the image masks used to create the point clouds. This similarity is scaled to be between 0 and 1. It can be formulated as:
\begin{equation}
\text{Semantic Similarity} = \frac{1}{2} \left( 1 + \frac{\mathbf{f}_{pt_{i}} \cdot \mathbf{f}_{pt_{j}}}{\|\mathbf{f}_{pt_{i}}\| \|\mathbf{f}_{pt_{j}}\|} \right)
\end{equation}
where $\mathbf{f}_{pt_{i}}$ and $\mathbf{f}_{pt_{j}}$ are the DINO feature vectors of the image masks corresponding to point clouds $pt_{i}$ and $pt_{j}$, respectively, $\cdot$ denotes the dot product, and $\|\cdot\|$ represents the norm of the vector.

Merging involves the integration of point clouds and the averaging of feature embeddings. A weighted average of CLIP and DINO embeddings is calculated, considering the contribution from the source key information, with a preference for nodes with more source identifiers. If a new node needs to be added, it is incorporated into the scene dictionary.

Scene refinement occurs once objects from all images in the input sequence have been added. This process consolidates nodes that represent the same physical objects but were initially identified as separate due to occlusions, viewpoint changes, or similar factors. It employs an overlap matrix to identify nodes that share spatial occupancy and logically merges them into a single node. The scene is finalized by discarding nodes that fail to meet the minimum number of points or detection criteria. This results in a refined and optimized final scene representation - \textbf{OpenSet 3D Semantic Instance Maps, a.k.a., O3D-SIM}.

\subsection{Language-Guided Navigation} \label{sec:lang_guided}
In this section, we leverage the LLM-based approach from \cite{Nanwani_2023}, which uses ChatGPT 3.5~\cite{OpenAI} to understand and map language commands to pre-defined function primitives that the robot can understand and execute. We have utilized the ChatGPT 3.5 API for our project due to its user-friendly interface and to have an accurate and fair comparison with SI Maps~\cite{Nanwani_2023}, as it also uses ChatGPT. It's noteworthy that other large language models (LLMs), such as LLAMA~\cite{touvron2023llama}, can achieve comparable results when provided with appropriate context for this process..

\begin{figure}
    \centering
    \includegraphics[width=\linewidth]{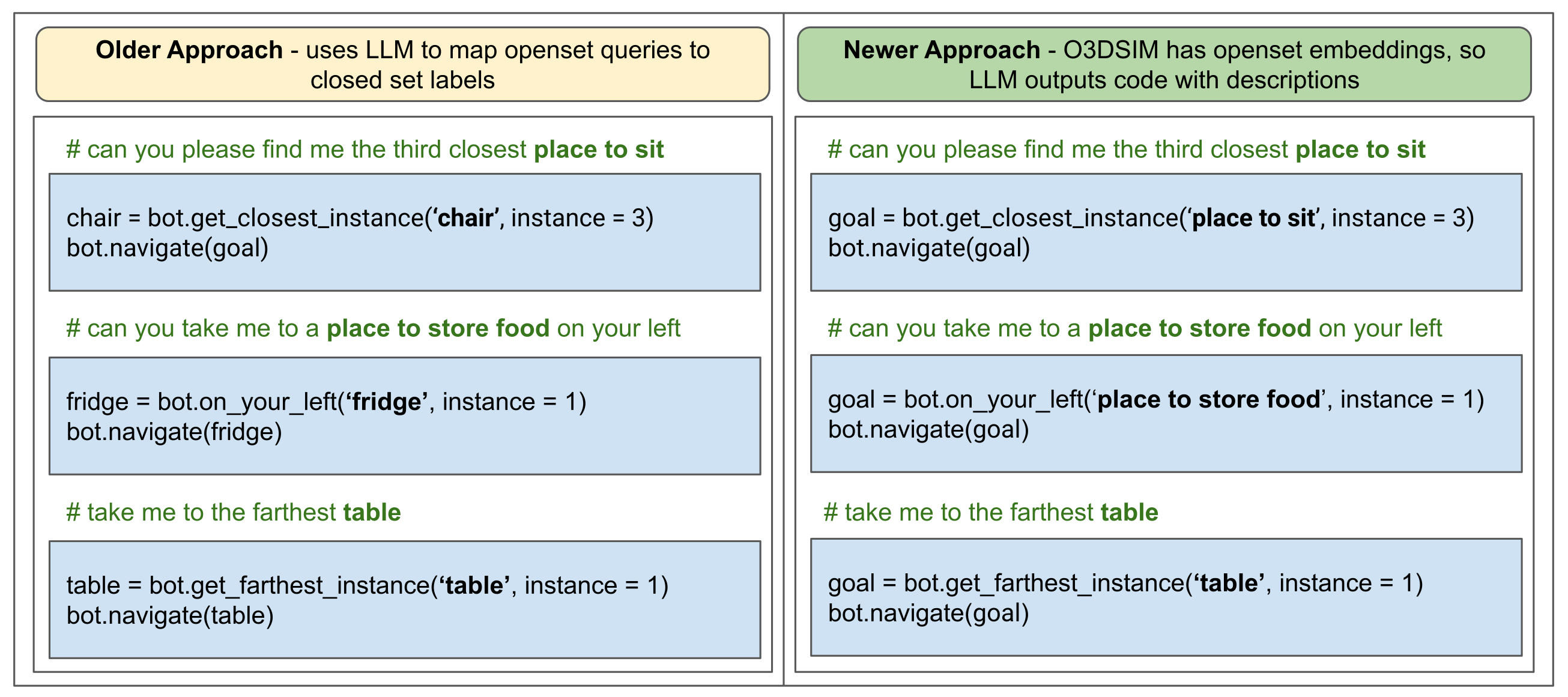}
    \caption{This figure shows the difference in output from ChatGPT due to the difference in nature of the two mapping approaches, where SI-Maps is closed-set, and O3D-SIM is open-set. For queries specifying exact object classes, both approaches output the same code. But, for queries specified in an open-set manner, the newer approach describes the goal to the code, whereas the older approach maps the description to the pre-known classes and passes this class to the code. The older approach benefits from LLM's understanding, whereas the newer approach benefits from the open-set embeddings (CLIP)}
    \label{fig:chatgpt}
\end{figure}

However, there are a few differences between our current approach and the approach in \cite{Nanwani_2023} regarding the use case of the LLM and the implementation of our function primitives. The previous approach used the LLM's ability to bring in an open-set understanding by mapping general queries to the already-known closed-set class labels obtained via Mask2Former~\cite{cheng2021mask2former}. However, given the open-set nature of our new representation, O3D-SIM, the LLM does not need to do that. Figure ~\ref{fig:chatgpt} shows both approaches' code output differences. The function primitives work similarly to the older approach, requiring the desired object type and its instance as an input. But now, the desired object is not from a pre-defined set of classes but a small query defining the object, so the implementation to find the desired location changes. We use the text and image-aligned nature of CLIP embeddings to find the desired object, where the input description is passed to the model, and its corresponding embedding is used to find the object in O3D-SIM. A cosine similarity is calculated between the embedding of the description and all the embeddings of our representation. These are ranked in an order based on the input query, and the desired instance is selected. This helps localise the required object instance from the language query.
%Once the instance is finalized, a goal corresponding to this instance is generated and passed to the navigation stack for autonomous navigation of the robot, hence achieving Language-Guided Navigation.

\begin{figure}
    \centering
    \includegraphics[width=\linewidth]{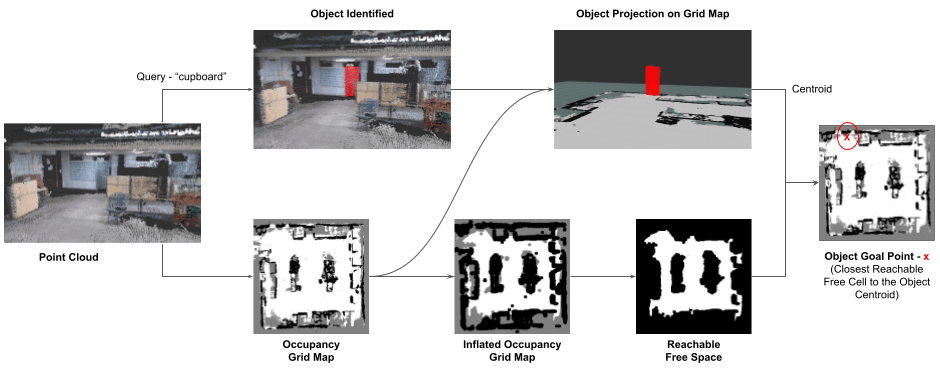}
    \caption{This figure shows the process of generating a goal point once the target object has been localised based on a sample language query and O3D-SIM.}
    \label{fig:goal_generation}
\end{figure}

Once the object and the desired instance have been localised using the language command, a goal point is generated so the robot can perform autonomous navigation towards this object. We create a top-down occupancy grid map from the point cloud generated from RGB-D data and poses. Since the object is represented as a point cloud, we have multiple points for a single instance; hence, we calculate the object's centroid. The obstacles in the occupancy map are inflated by the robot's radius, and reachable free space is marked in this inflated map using Breadth-First Search (BFS). Finally, the object centroid is projected onto the grid map, and the closest reachable free grid cell to the centroid is returned to the robot as the goal point for autonomous navigation (Figure~\ref{fig:goal_generation}), therefore achieving Language-Guided Navigation.

\section{Experiments}\label{sec:experiments}

Having introduced the O3D-SIM creation pipeline and its integration with ChatGPT for natural language understanding and Vision-Language Navigation (VLN) enhancement, we now turn to the evaluation of this novel representation both quantitatively and qualitatively. This will also shed light on the impact of the O3D-SIM representation on an agent's ability to execute queries that mimic human interaction. Figure \ref{fig:figure1} shows a few examples of such queries. The evaluation is structured into two subsections: Section \ref{sec:quantitative} focuses on the quantitative assessment of O3D-SIM, and Section \ref{sec:qualitative} addresses the qualitative analysis of the representation. 
The section \ref{sec:setup} details our hardware configurations for creating and testing the maps during real-world experiments.

\subsection{Hardware Setup}\label{sec:setup}
Creation of O3D-SIM is an offline pipeline and leverages large foundational models that require a good amount of GPU VRAM. Hence, we use a server setup with a 16 GB Nvidia RTX A4000, 64 GB RAM and an AMD Ryzen 7 3800X 8-core processor to create the maps. For collecting data in the real world and testing out the navigation performance, we use a custom robot we have built in our lab, which provides wheel odometry using encoders, planar lidar scans using an RPLiDAR sensor and RGB-D data using a Kinect Azure Camera. For inference on the robot, we use a laptop with a 6 GB Nvidia RTX 3060, 32 GB RAM and an AMD Ryzen 7 5800H 8-core processor. We load O3D-SIM on the laptop and integrate it with the ROS Navigation Stack, where the goal is generated from language commands and passed to the mobile robot (section \ref{sec:lang_guided}). We do not require the GPU on the laptop to infer from the CLIP model and O3D-SIM, and the object localisation using O3D-SIM and CLIP and the corresponding goal point generation happens in about a second. The inference time for the language query to code generation depends on the response time of the LLM's API. It took an average time of about 45-60 seconds with ChatGPT 3.5 for our experiments, which is consistent with the time reported in other works.

\subsection{Quantitative Evaluation}\label{sec:quantitative}

To facilitate the construction of O3D-SIM and its quantitative evaluation, we employ the Matterport3D dataset~\cite{chang2017matterport3d} within the Habitat simulator~\cite{savva2019habitat}. Matterport3D, a comprehensive RGB-D dataset, encompasses 10,800 panoramic views derived from 194,400 RGB-D images across 90 large-scale buildings. It offers surface reconstructions, camera poses, and 2D and 3D semantic segmentations — critical components for creating accurate Ground-truth models. Both Matterport3D and Habitat are widely utilized for assessing the navigational abilities of VLN agents in indoor settings, enabling robots to execute navigational tasks dictated by natural language commands in a seamless environment, with performance meticulously documented. To evaluate O3D-SIM, we compiled 5,267 RGB-D frames and their respective pose data from five distinct scenes, applying this dataset across all mapping pipelines included in our assessment. We also gathered real-world environment data for evaluation purposes, expanding our analysis to encompass six unique scenes.

\textbf{Baseline:} We evaluate the performance of the O3D-SIM against the logical baseline used in our previous work, VLMaps with Connected Components, and also evaluate against our approach SI Maps from \cite{Nanwani_2023}. The three methods mentioned for comparison are chosen as they try to achieve things similar to our approach.

\textbf{Evaluation Metrics:} Like prior approaches \cite{huang23vlmaps, schumann-riezler-2022-analyzing, jain2022ground} in VLN literature, we use the gold standard \textbf{\textit{Success Rate}} metric, also known as \textbf{\textit{Task Completion}} metric to measure the success ratio for the navigation task. We choose \emph{Success Rate} on navigation tasks as they directly quantify the overall approach and indirectly quantify the performance of O3D-SIM in detecting the instances because if the instances along the way are not properly detected, the queries are bound to fail. We compute the \emph{Success Rate} metric through human and automatic evaluations. For automatic evaluation, we define success if the agent reaches within a threshold distance of the ground truth goal. Here, the agent's orientation concerning the goal(s) doesn't matter and might show success even when the agent fails. For example, if there are multiple paintings and the agent is asked to point to a particular painting at the end of a query, the agent may reach with a close distance of the desired painting but end up looking at something undesired. Hence, we also use human evaluation to verify if the agent ends up in a desired position according to the query. Human Verification takes in votes from the three people and decides, based on these votes, the success of a task.

\begin{figure}
    \centering
    \includegraphics[width=\linewidth]{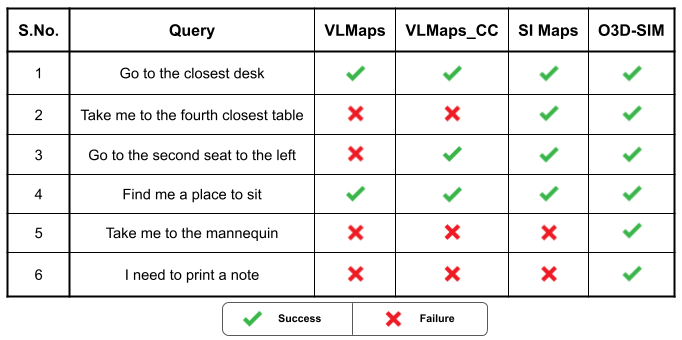}
    \caption{The figure shows the success/failure of the four representations in our lab environment for a few open/closed-set queries.}
    \label{fig:queries_example}
\end{figure}

\textbf{Results:} We present the results of the evaluation metric \emph{Success Rate} in Table~\ref{tab:success_rate}. In our experimentation, we observe a remarkable improvement in performance compared to the other approaches we have shown in our paper, especially against the baselines from \cite{Nanwani_2023}. O3D-SIM performs better than VLMaps with CC due to its ability to identify instances robustly. SI Maps and O3D-SIM perform better than the baselines due to their ability to separate instances. However, O3D-SIM has the edge over SI Maps due to its open set and 3D nature, allowing it to understand the surroundings better. 
Figure~\ref{fig:queries_example} shows a few language queries we tested in our lab using all four representations - VLMaps, VLMaps\_CC, SI Maps and O3D-SIM.

\begin{table}%[h]
\centering
\renewcommand{\arraystretch}{1.5}
\begin{tabular}{lcc}
\hline
{Method} & \multicolumn{2}{c}{Success Rate} \\ \cline{2-3} 
& Human Evaluation & Automatic Evaluation 
\\ 
\hline
VL Maps                 & 0.40               & 0.46               \\
VL Maps with CC         & 0.48               & 0.52               \\
SI Maps (K=5)           & 0.70               & 0.74      
\\
SI Maps (K=9)           & 0.72               & 0.74       
\\
O3D-SIM                 & \textbf{0.82}      & \textbf{0.84}
\\
\hline
\\
\end{tabular}
\caption{O3D-SIM outperforms the baseline methods from \cite{Nanwani_2023} by significantly large margins on the \emph{Success Rate} metric. It also shows an improvement from our previous approach, i.e., SI-Maps. The best results are highlighted in \textbf{bold}. For this evaluation, the agent executes a set of open-set and closed-set queries in 5 different scenes from Matterport3D and 1 real world environment.}
\label{tab:success_rate}
\end{table}

\subsection{Qualitative Results}\label{sec:qualitative}
\begin{figure}
    \centering
    \includegraphics[width=0.9\linewidth]{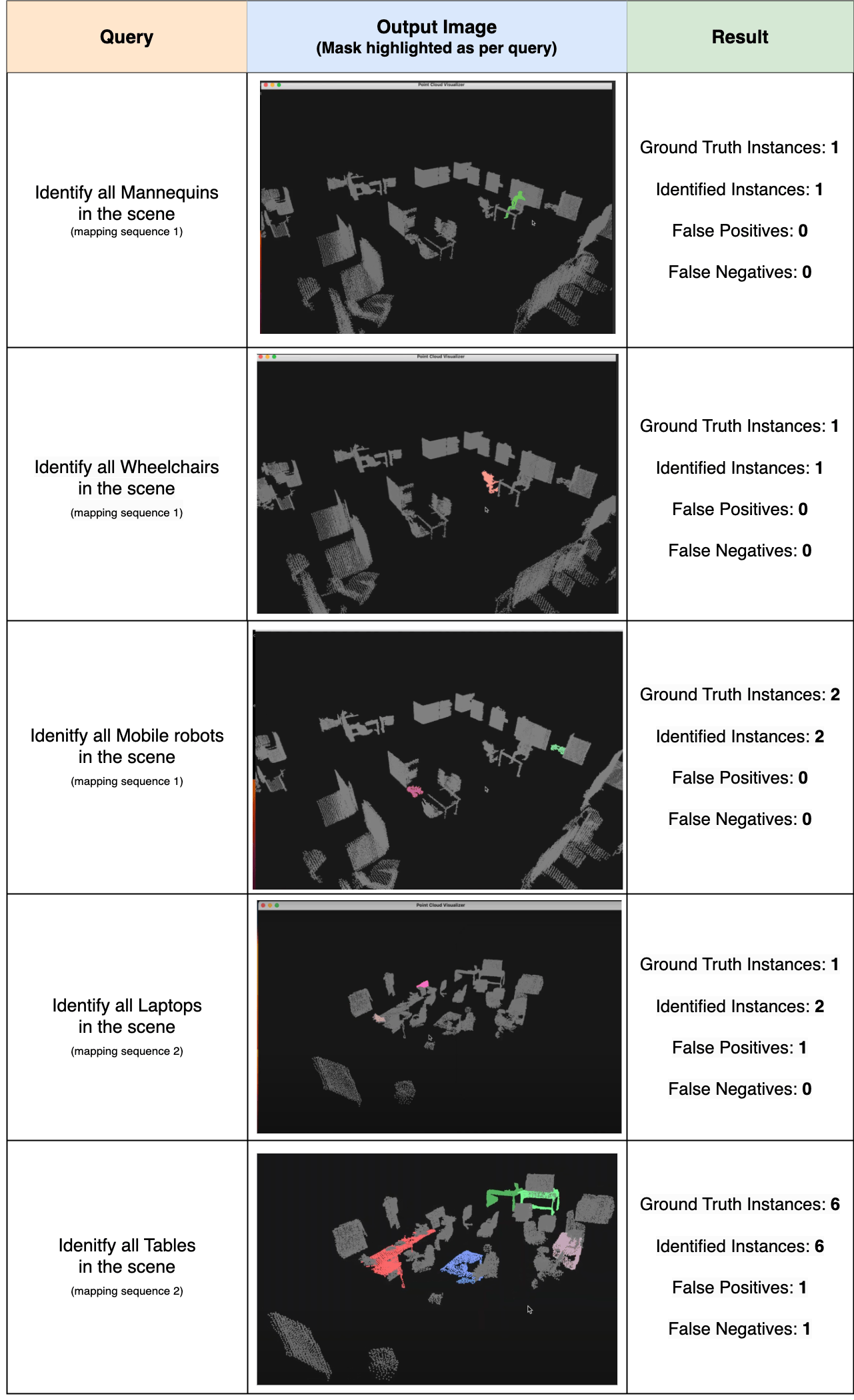}
   \caption{This figure presents the qualitative results for O3D-SIM. Mapping sequence 1 was recorded using a Kinect Azure on a mobile robot, whereas mapping sequence 2 was recorded using an iPad, both in the real world. The environment was changed between the two sequences. The first four queries show some unique objects identified by O3D-SIM that normal closed-set approaches may fail to identify.}
  \label{fig:qualitative}
  \vspace{-1em}
\end{figure}

To qualitatively demonstrate the effectiveness of the proposed O3D-SIM, this section includes visualizations of the model's performance using select examples. These visualizations are displayed in Figure \ref{fig:qualitative}, illustrating the outcomes for two mapping sequences. 

Notably, the open-set capability of O3D-SIM enables the identification of objects that are typically undetectable by conventional pipelines relying on closed sets or predefined datasets, such as wheelchairs. The figure showcases a comparative analysis of our pipeline's ability to accurately identify and segment various objects, including mannequins and mobile robots, against their actual counts. This comparison highlights situations where traditional methods, constrained by a limited set of recognizable objects, fall short.

Our approach recognises instance-level semantics, accurately identifying 5 out of 6 table instances (with one false positive), underscoring its precision across both simulated and real-world data. This demonstrates the robustness of our pipeline, further evidenced by the clarity of the semantic map and the ease with which instance-level segmentation results can be visualized. While methods like VLMaps might identify a broader range of objects due to their open-set nature, and SI-Maps may detect multiple instances of the same object, O3D-SIM uniquely excels at both, offering a comprehensive solution.

\section{Conclusion and Future Work}\label{sec:futureworkconclusion}

In conclusion, we introduce O3D-SIM, an innovative approach to scene representation that builds on the concepts established in our previous work, SI-Maps. We demonstrated that an instance-level understanding of an environment enhances performance in language-guided tasks. O3D-SIM advances the 2D closed-set approach of SI-Maps to an open-set 3D representation. When combined with a Large Language Model (LLM) to incorporate natural language understanding, O3D-SIM significantly improves the success rate of executing natural language queries. This advancement suggests that representations akin to O3D-SIM could enable robots to approach the comprehension and execution of natural language commands nearly as effectively as humans.

A promising avenue for future development involves integrating dynamic objects into O3D-SIM. Although it currently identifies static objects, equipping the pipeline to recognize humans and other dynamic entities would enable handling more complex queries, bringing them closer to human levels of understanding. Such an enhancement would be invaluable in various settings, including hospitals and household robotics, or any context where human-robot collaboration is crucial.
Another intriguing avenue for exploration is the integration of 3D instance-level scene representations, such as O3D-SIM, with physics engines to transform real-life environments into life-like simulations, akin to Nvidia Isaac Gym~\cite{makoviychuk2021isaac}. Given O3D-SIM's capability to distinguish between objects at an instance level, these simulations could be dynamically modified as needed without the necessity of re-recording the entire scene. This flexibility allows for creating varied iterations of the same life-like environment.

\section*{Disclosure statement}
No potential conflict of interest was reported by the author(s).

\section*{Acknowledgements}
The author, Laksh Nanwani, thanks IHub-Data, IIIT Hyderabad, for
extending their research fellowship. We also acknowledge IHub-Data for supporting this work.

\section*{Notes on contributors}

\textbf{\emph{Laksh Nanwani}} is pursuing his PhD at IIIT Hyderabad, India, and received his B.E. in Computer Science from BITS Pilani, India, in 2020. He is associated with the Robotics Research Center at IIIT Hyderabad, where he is working on a Smart Autonomous Wheelchair project under the guidance of Prof. K Madhava Krishna. His research encompasses Robotics, Computer Vision, and Planning.
\\~\\
\textbf{\emph{Kumaraditya Gupta}} is a Research Associate at the Robotics Research Center, IIIT Hyderabad, India. He holds a B.E. in Electrical and Electronics Engineering from BITS Pilani, India. His research interests lie at the intersection of Robotics and Computer Vision.
\\~\\
\textbf{\emph{Aditya Mathur}} is a Research Associate at the Robotics Research Center, IIIT Hyderabad, India. He holds a B.Tech in Mechanical Engineering from G.L. Bajaj Institute of Technology, India. He has gained valuable experience in both corporate and research roles, contributing to his expertise in the field. His research interests focus on the intersection of Robotics and Computer Vision.
\\~\\
\textbf{\emph{Swayam Agrawal}} is an undergraduate student pursuing Computer Science with Honours at IIIT Hyderabad. He conducts research at the Robotics Research Center as part of his coursework, and his interests lie in software systems, robotics, and 3D vision, with current research focusing on scene reconstruction and relative pose estimation and working with foundation models.
\\~\\
\textbf{\emph{A.H. Abdul Hafez}}, PhD, FHEA, is an Associate Professor at Hasan Kalyoncu University, Gaziantep, Turkey. He earned his PhD in Computer Science and Engineering in 2008 from Osmania University in collaboration with IIIT Hyderabad, India. He is a Fellow of the Higher Education Academy (FHEA). Abdul Hafez has published extensively in prestigious conferences and journals such as ICRA and IROS. His current research interests include robotic vision, machine learning, and autonomous and assistive technologies.
\\~\\
\textbf{\emph{K Madhava Krishna}}, FNAE (Fellow of the National Academy of Engineers, India), is a Professor at IIIT Hyderabad, India and heads the Robotics Research Center there. The Center is ranked first in the country's top 6 in Asia and among the top 40 globally, according to csrankings.org and airankings.org. Krishna has more than 80 publications in flagship robotic conferences (ICRA, IROS, and RAL) and contributes to the areas of robotic vision, SLAM, perception, coupled navigation for ground and aerial robots, and self-driving cars.

\bibliography{references}

\end{document}